\documentclass[11pt]{article}

\usepackage[utf8]{inputenc}
\usepackage[T1]{fontenc}
\usepackage{amsmath,amssymb,amsfonts}
\usepackage{graphicx}
\usepackage{booktabs}
\usepackage{hyperref}
\usepackage{xcolor}
\usepackage{algorithm}
\usepackage{algorithmic}
\usepackage[margin=1in]{geometry}
\usepackage{caption}
\usepackage{subcaption}
\usepackage{float}

\graphicspath{{figures/}}

\newcommand{\tacit}{\textsc{Tacit}}
\newcommand{\R}{\mathbb{R}}
\newcommand{\E}{\mathbb{E}}

\hypersetup{
    colorlinks=true,
    linkcolor=blue,
    citecolor=blue,
    urlcolor=blue
}

\title{TACIT: Transformation-Aware Capturing of Implicit Thought\\
\Large Flow Matching for Interpretable Visual Reasoning}

\author{
    Daniel Nobrega Medeiros\\
    Independent Researcher\\
    \texttt{ORCID: 0000-0003-3604-7380}\\
    \url{https://github.com/danielxmed/tacit}
}

\date{}

\begin{document}

\maketitle

\begin{abstract}

We present \tacit{} (Transformation-Aware Capturing of Implicit Thought), a diffusion-based transformer model designed for interpretable visual reasoning. Our core hypothesis is that flow matching between visual states can capture structural transformations---a form of ``visual intuition''---without relying on language. Unlike language-based reasoning systems, \tacit{} operates entirely in pixel space, enabling direct visualization of the reasoning process at each inference step. We demonstrate the approach on maze-solving, where the model learns to transform images of unsolved mazes into their solutions. Using rectified flow with deterministic Euler sampling, intermediate states reveal an interpretable progression from problem to solution. Trained on 1 million synthetic maze pairs, our model achieves a 192$\times$ reduction in training loss and produces visually accurate solutions. Quantitative analysis reveals a striking phase transition: the solution remains invisible for 68\% of the transformation, then emerges abruptly and simultaneously across the entire path within just 2\% of the process---a pattern reminiscent of ``eureka'' moments in human insight. The pixel-space design, combined with noise-free flow matching, provides a foundation for understanding how neural networks develop and apply implicit reasoning strategies. Code, model weights, and dataset are publicly available.

\end{abstract}

\section{Introduction}
\label{sec:introduction}

Before we articulate a solution, we have often already \textit{seen} it---in a flash of insight, a felt sense, a wordless understanding. The chess grandmaster perceives the winning position before calculating the moves. The experienced engineer glances at a design and knows something is wrong. The mathematician sees a proof's structure before formalizing its steps. This capacity for pre-linguistic understanding---what philosopher Michael Polanyi termed ``tacit knowledge'' \cite{polanyi1966tacit}---represents a fundamental aspect of intelligence that resists explicit articulation.

While recent advances in large language models have demonstrated impressive reasoning capabilities through chain-of-thought prompting \cite{wei2022chain}, these approaches fundamentally rely on linguistic scaffolding. Reasoning is externalized into words; understanding is translated into text. This works remarkably well for many tasks, but it may not capture the full range of human cognitive capabilities---particularly those visual-spatial intuitions that operate below and before language. Cognitive science distinguishes between ``System 1'' (fast, intuitive, automatic) and ``System 2'' (slow, deliberate, verbal) thinking \cite{kahneman2011thinking}; chain-of-thought methods externalize System 2 processes, but the rapid pattern recognition of System 1 remains largely unaddressed. Recent proposals for extending deep learning toward ``System 2'' capabilities \cite{bengio2019system2} suggest the need for architectures that can capture both modes of cognition.

We propose an alternative approach: learning visual reasoning through flow matching between problem and solution states, entirely bypassing language. This work introduces \tacit{} (Transformation-Aware Capturing of Implicit Thought), a diffusion-based transformer that learns to capture structural transformations in a purely visual domain. Our central hypothesis is:

\begin{quote}
\textit{Flow matching from problem state $t=0$ to solution state $t=1$ can learn structural transformations in a language-independent manner, with intermediate states ($0 < t < 1$) encoding interpretable reasoning steps.}
\end{quote}

To test this hypothesis, we focus on maze-solving as our reasoning domain. Mazes provide an ideal testbed because: (1) they require genuine reasoning---finding a valid path from entrance to exit; (2) the problem-solution relationship has clear visual structure; (3) solutions are verifiable; and (4) the domain is simple enough to enable detailed analysis while being complex enough to require non-trivial computation.

Our approach differs from prior work in several key design decisions that prioritize interpretability:

\begin{itemize}
    \item \textbf{Pixel-space operation}: Unlike most diffusion models that use latent-space encodings (VAE), we operate directly on pixels. This preserves discrete structural information crucial for logical reasoning and ensures that intermediate inference states are actual images that can be directly visualized.

    \item \textbf{Rectified flow}: We use rectified flow (flow matching) rather than DDPM-style diffusion. This eliminates stochastic noise injection, making the transformation deterministic and the intermediate states meaningful rather than noisy.

    \item \textbf{Euler sampling}: Our inference uses simple Euler integration with 10 steps. Each step produces a valid image, allowing us to observe the model's ``thought process'' as it transforms the problem into the solution.
\end{itemize}

The resulting system learns to solve mazes with high accuracy while providing interpretable intermediate states. We train on 1 million synthetic maze-solution pairs and demonstrate:

\begin{enumerate}
    \item Successful learning of maze-solving, with 192$\times$ reduction in training loss over 100 epochs
    \item Qualitatively interpretable inference trajectories showing progressive solution construction
    \item A 22.7$\times$ improvement in prediction quality (L2 distance to ground truth)
    \item A sharp phase transition at $t = 0.70$ where solutions emerge within 2\% of the transformation time
    \item 100\% simultaneous emergence across spatial regions, ruling out sequential construction
\end{enumerate}

Beyond the immediate results, this work establishes a framework for studying implicit reasoning in neural networks. By removing language from the loop and operating in interpretable pixel space, we can directly examine how models develop internal representations of problem-solving strategies---representations that may parallel the tacit knowledge humans possess but cannot easily articulate.

\section{Related Work}
\label{sec:related}

\paragraph{Diffusion Models and Flow Matching.}
Diffusion models have emerged as powerful generative models, with denoising diffusion probabilistic models (DDPM) \cite{ho2020ddpm} establishing the foundation. Flow matching \cite{lipman2023flow} and rectified flow \cite{liu2023flow} provide an alternative formulation that learns direct transformations between distributions without the noise injection characteristic of DDPM. Our work uses rectified flow for its deterministic properties, which enable interpretable intermediate states during inference. The recent success of Stable Diffusion 3 \cite{esser2024sd3}, which employs rectified flow transformers at scale, validates the effectiveness of this approach for high-quality image generation.

\paragraph{Vision Transformers and DiT.}
The Vision Transformer (ViT) \cite{dosovitskiy2021vit} demonstrated that pure transformer architectures can achieve state-of-the-art results on image tasks, building on the attention mechanism introduced in \cite{vaswani2017attention}. The Diffusion Transformer (DiT) \cite{peebles2023dit} extended this to generative modeling, replacing the U-Net backbone common in diffusion models with a transformer architecture. Our architecture builds on DiT, using adaptive layer normalization (adaLN) for timestep conditioning and patch-based image tokenization.

\paragraph{Visual Reasoning.}
Visual reasoning has been studied through various benchmarks including RAVEN \cite{zhang2019raven}, PGM \cite{barrett2018pgm}, and ARC \cite{chollet2019arc}. These typically evaluate models on abstract reasoning patterns. Our maze-solving task represents a complementary approach focused on spatial reasoning and path planning, with the advantage of straightforward solution verification.

\paragraph{World Models and Latent Representations.}
Recent work on world models \cite{ha2018world} explores how neural networks can learn internal representations of environment dynamics. JEPA (Joint Embedding Predictive Architecture) \cite{lecun2022path} proposes learning in latent space to capture abstract features. Our work takes an opposite approach---operating in pixel space to maximize interpretability---while sharing the goal of understanding implicit knowledge representation.

\paragraph{Interpretability in Neural Networks.}
Mechanistic interpretability aims to understand the internal computations of neural networks. Our design choices (pixel space, rectified flow, deterministic sampling) are motivated by interpretability: we want to observe the reasoning process, not just the final output.

\section{Method}
\label{sec:method}

\subsection{Problem Formulation}

We formulate visual reasoning as a conditional transformation problem. Given an input image $x_0$ representing an unsolved problem (maze without solution path), we aim to generate $x_1$, the corresponding solution (maze with solution path marked). The model learns this transformation through flow matching.

\subsection{Flow Matching Objective}

Following rectified flow \cite{liu2023flow}, we define a linear interpolation between problem and solution:
\begin{equation}
    x_t = (1 - t) \cdot x_0 + t \cdot x_1, \quad t \in [0, 1]
\end{equation}
where $x_0$ is the input image and $x_1$ is the target solution. The target velocity field is simply the direction from problem to solution:
\begin{equation}
    v_{\text{target}} = x_1 - x_0
\end{equation}

The model $f_\theta$ learns to predict this velocity field given the current state and timestep:
\begin{equation}
    v_{\text{pred}} = f_\theta(x_t, t)
\end{equation}

Training minimizes the mean squared error between predicted and target velocities:
\begin{equation}
    \mathcal{L} = \E_{t \sim \mathcal{U}(0,1)} \left[ \| f_\theta(x_t, t) - v_{\text{target}} \|^2 \right]
\end{equation}

This formulation has several advantages for interpretability:
\begin{itemize}
    \item No noise injection: Unlike DDPM, the interpolated state $x_t$ is a clean blend of input and output
    \item Linear paths: The optimal transport is along straight lines in pixel space
    \item Deterministic inference: Sampling follows the learned velocity field without stochasticity
\end{itemize}

\paragraph{Philosophical significance of noise-free transport.}
This is not merely a technical convenience. Traditional diffusion models map between a noise distribution and a data distribution:
\begin{equation}
    p_{\text{noise}} \rightarrow p_{\text{data}}
\end{equation}
The intermediate states in such models are corrupted by noise---they carry no clear semantic content and are mathematical conveniences rather than meaningful representations.

In contrast, our flow matching formulation maps between two \textit{meaningful} distributions:
\begin{equation}
    p_{\text{problem}} \rightarrow p_{\text{solution}}
\end{equation}
Both endpoints carry structured, interpretable information. The flow between them is not a denoising process---it is a \textit{transformation process}. The intermediate state $x_{t=0.5}$ is the midpoint of a meaningful trajectory between problem and solution: not yet a solution, no longer just a problem, but a transitional state that carries the structure of the transformation.

This makes our intermediate representations interpretable in principle. When we observe $x_t$ at various values of $t$, we are observing the model's ``thought in progress''---its implicit computation made visible. The model, in a sense, \textit{dreams} its way from problem to solution, and we can watch the dream unfold.

\subsection{Model Architecture}

\begin{figure}[t]
\centering
\includegraphics[width=0.95\textwidth]{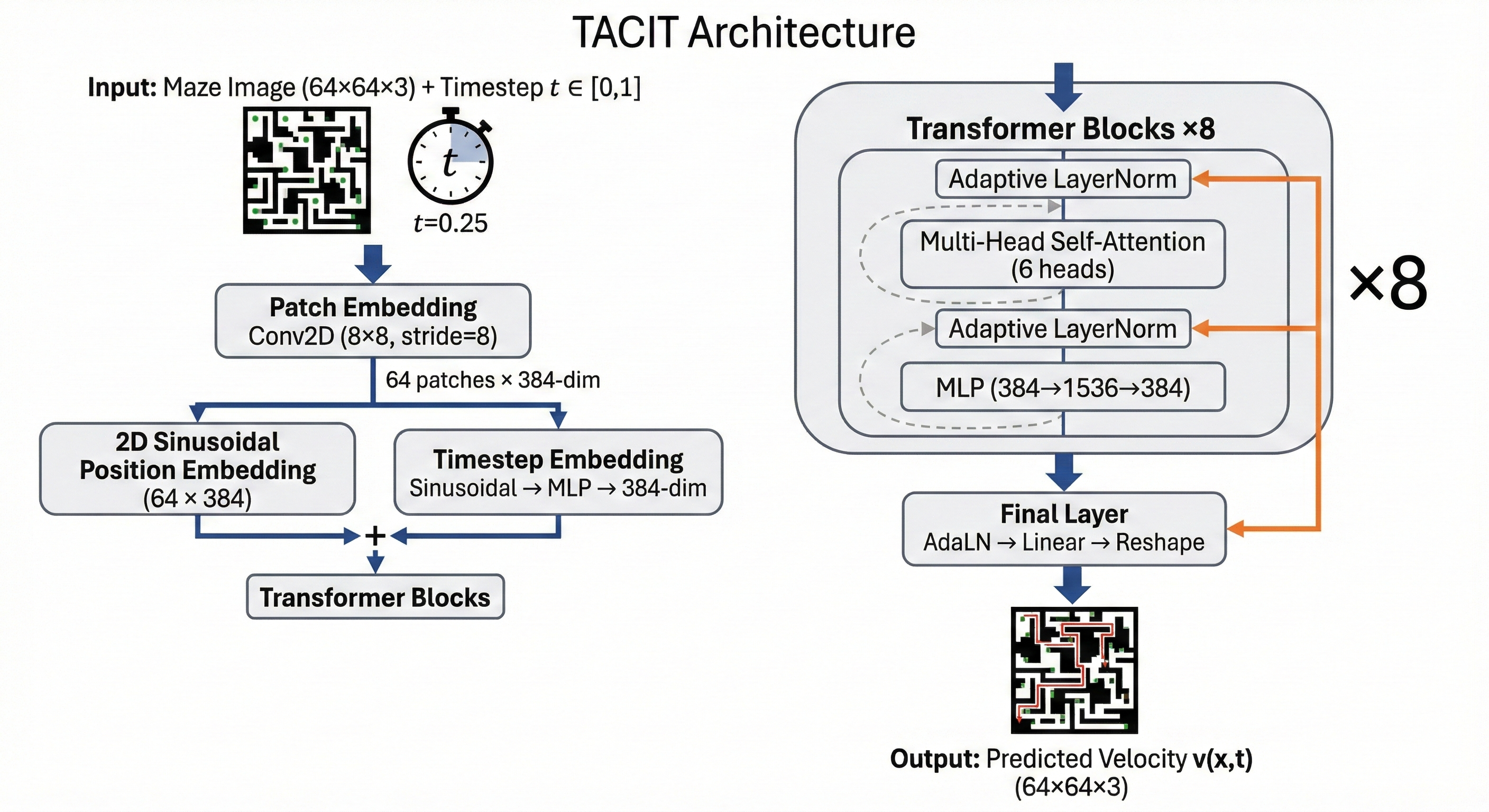}
\caption{\tacit{} architecture overview. Input images (64$\times$64$\times$3) are tokenized into 64 patches via convolutional embedding, enriched with 2D sinusoidal positional encodings, and processed through 8 transformer blocks with adaptive layer normalization (adaLN) conditioned on the timestep embedding. The final layer reconstructs the predicted velocity field in pixel space. Orange arrows indicate timestep conditioning; the model learns to predict the direction from problem to solution at each point along the interpolation path.}
\label{fig:architecture}
\end{figure}

\tacit{} uses a Diffusion Transformer (DiT) architecture adapted for image-to-image reasoning. The architecture follows a standard DiT design with modifications for image-to-image transformation. Figure~\ref{fig:architecture} provides an overview of the complete architecture.

\subsubsection{Patch Embedding}

Input images of size $64 \times 64 \times 3$ are divided into non-overlapping $8 \times 8$ patches, resulting in 64 patch tokens. Each patch is linearly projected to the hidden dimension:
\begin{equation}
    z_i = W_{\text{patch}} \cdot \text{flatten}(\text{patch}_i) + b_{\text{patch}}, \quad z_i \in \R^{384}
\end{equation}

The patch embedding is implemented as a 2D convolution with kernel size and stride equal to the patch size (8), effectively performing the flattening and projection in one operation.

\subsubsection{Positional Encoding}

We use 2D sinusoidal positional embeddings to encode spatial information. For each patch position $(p_x, p_y)$ on the $8 \times 8$ grid:
\begin{align}
    \text{PE}_{(p_x, p_y, 2i)} &= \sin\left(\frac{p_x}{10000^{2i/d}}\right) \\
    \text{PE}_{(p_x, p_y, 2i+1)} &= \cos\left(\frac{p_x}{10000^{2i/d}}\right)
\end{align}
with similar terms for $p_y$. The x and y embeddings are concatenated to form the full 384-dimensional position embedding, which is added to the patch embeddings.

\subsubsection{Timestep Embedding}

The diffusion timestep $t \in [0, 1]$ is encoded using sinusoidal embeddings followed by a two-layer MLP:
\begin{equation}
    e_t = \text{MLP}(\text{SinusoidalEmbed}(t)) \in \R^{384}
\end{equation}

The sinusoidal embedding uses 256 frequency components, providing multiple ``perspectives'' on the timestep at different temporal resolutions. The MLP consists of two linear layers (256 $\to$ 384 $\to$ 384) with SiLU activation.

\subsubsection{Transformer Blocks with Adaptive Layer Normalization}

The model consists of 8 transformer blocks. Each block uses \textit{adaptive layer normalization} (adaLN) to condition on the timestep embedding. Unlike standard layer normalization, adaLN learns to modulate the normalized activations based on the conditioning signal:
\begin{equation}
    \text{adaLN}(h, e_t) = \gamma(e_t) \odot \text{LayerNorm}(h) + \beta(e_t)
\end{equation}
where $\gamma$ and $\beta$ are predicted from the timestep embedding via a linear layer.

Each transformer block contains:
\begin{enumerate}
    \item \textbf{Adaptive LayerNorm 1}: Conditions the input before attention
    \item \textbf{Multi-Head Self-Attention}: 6 attention heads with head dimension 64
    \item \textbf{Residual connection}
    \item \textbf{Adaptive LayerNorm 2}: Conditions before the feedforward network
    \item \textbf{Feedforward Network}: Two-layer MLP (384 $\to$ 1536 $\to$ 384) with GELU activation
    \item \textbf{Residual connection}
\end{enumerate}

The self-attention mechanism allows each patch to attend to all other patches:
\begin{equation}
    \text{Attention}(Q, K, V) = \text{softmax}\left(\frac{QK^T}{\sqrt{d_k}}\right) V
\end{equation}
where $Q$, $K$, $V$ are linear projections of the input and $d_k = 64$ is the head dimension. We use PyTorch's \texttt{scaled\_dot\_product\_attention} which automatically selects efficient implementations (Flash Attention when available).

\subsubsection{Final Layer}

The final layer converts patch tokens back to image pixels:
\begin{enumerate}
    \item Adaptive layer normalization conditioned on timestep
    \item Linear projection: 384 $\to$ 192 (= $8 \times 8 \times 3$ pixels per patch)
    \item Reshape to image format: $(B, 64, 192) \to (B, 3, 64, 64)$
\end{enumerate}

\subsubsection{Architecture Summary}

Table~\ref{tab:architecture} summarizes the model hyperparameters.

\begin{table}[h]
\centering
\caption{Model architecture hyperparameters}
\label{tab:architecture}
\begin{tabular}{@{}ll@{}}
\toprule
\textbf{Component} & \textbf{Specification} \\
\midrule
Input resolution & $64 \times 64 \times 3$ \\
Patch size & $8 \times 8$ \\
Number of patches & 64 \\
Hidden dimension ($d$) & 384 \\
Transformer blocks & 8 \\
Attention heads & 6 \\
Head dimension & 64 \\
MLP hidden dimension & 1536 (4$\times d$) \\
Timestep embedding dimension & 256 \\
Total parameters & $\sim$20M \\
\bottomrule
\end{tabular}
\end{table}

\subsection{Inference via Euler Sampling}

At inference time, we integrate the learned velocity field using the Euler method:
\begin{equation}
    x_{t+\Delta t} = x_t + f_\theta(x_t, t) \cdot \Delta t
\end{equation}

Starting from $x_0$ (the unsolved maze), we take 10 Euler steps with $\Delta t = 0.1$ to arrive at the predicted solution $\hat{x}_1$. Algorithm~\ref{alg:euler} describes the sampling procedure.

\begin{algorithm}[h]
\caption{Euler Sampling for \tacit{}}
\label{alg:euler}
\begin{algorithmic}[1]
\REQUIRE Input image $x_0$, model $f_\theta$, number of steps $N=10$
\ENSURE Predicted solution $\hat{x}_1$
\STATE $x \leftarrow x_0$
\STATE $\Delta t \leftarrow 1/N$
\FOR{$i = 0$ to $N-1$}
    \STATE $t \leftarrow i \cdot \Delta t$
    \STATE $v \leftarrow f_\theta(x, t)$ \COMMENT{Predict velocity}
    \STATE $x \leftarrow x + v \cdot \Delta t$ \COMMENT{Euler step}
\ENDFOR
\RETURN $\text{clip}(x, 0, 1)$
\end{algorithmic}
\end{algorithm}

The deterministic nature of Euler sampling (no noise injection) means that each intermediate state $x_t$ is a meaningful interpolation that can be visualized. This is a key feature for interpretability: we can observe the model's ``reasoning trajectory'' from problem to solution.

Several properties of this inference procedure are noteworthy:
\begin{itemize}
    \item \textbf{Remarkably few steps}: Where DDPM-style diffusion typically requires 100-1000 denoising steps, our flow matching requires only 10 Euler steps. Each step is a ``thought snapshot''---a complete image representing the model's current state of transformation.

    \item \textbf{Determinism enables analysis}: The same input always produces the same trajectory. This reproducibility enables systematic analysis of intermediate states across different mazes, epochs, or model variants.

    \item \textbf{Computational efficiency}: Only 10 forward passes per sample, compared to hundreds or thousands for typical diffusion models. This efficiency is not incidental---it reflects the simpler structure of problem$\to$solution transport compared to noise$\to$data transport.
\end{itemize}

\section{Dataset}
\label{sec:dataset}

\subsection{Maze Generation}

We generate mazes using a randomized depth-first search (DFS) algorithm with iterative backtracking:

\begin{enumerate}
    \item Initialize a grid where all cells are walls
    \item Start from position (1, 1), mark as path, push to stack
    \item While stack is non-empty:
    \begin{enumerate}
        \item Get unvisited neighbors (2 cells away to maintain wall structure)
        \item If neighbors exist: randomly choose one, carve path, push to stack
        \item Otherwise: backtrack (pop from stack)
    \end{enumerate}
\end{enumerate}

This generates perfect mazes (exactly one path between any two points) with varied complexity.

\subsection{Solution Finding}

Solutions are found using breadth-first search (BFS) from the top-left entry point (1, 1) to the bottom-right exit (size-2, size-2). BFS guarantees the shortest path, which is marked in red on the solved maze image.

\subsection{Image Rendering}

Mazes are rendered as $64 \times 64$ RGB images using nearest-neighbor interpolation to preserve sharp edges:
\begin{itemize}
    \item \textbf{White} (255, 255, 255): Path cells (traversable)
    \item \textbf{Black} (0, 0, 0): Wall cells
    \item \textbf{Green} (0, 255, 0): Entry and exit points
    \item \textbf{Red} (255, 0, 0): Solution path (target only)
\end{itemize}

\subsection{Dataset Statistics}

We generate diverse mazes with varying logical grid sizes:

\begin{table}[h]
\centering
\caption{Dataset characteristics}
\label{tab:dataset}
\begin{tabular}{@{}ll@{}}
\toprule
\textbf{Property} & \textbf{Value} \\
\midrule
Total pairs & 1,000,000 \\
Maze sizes (logical grid) & 11, 15, 21, 25, 31 \\
Image resolution & $64 \times 64$ \\
Channels & 3 (RGB) \\
Storage format & Compressed NumPy (.npz) \\
Batch file size & $\sim$120 MB (10,000 samples) \\
Total storage & $\sim$12 GB \\
\bottomrule
\end{tabular}
\end{table}

The varying maze sizes ensure the model learns to handle different complexity levels. Smaller mazes (11$\times$11) have shorter solutions, while larger mazes (31$\times$31) require longer, more complex paths.

\subsection{Data Loading}

We implement lazy batch loading for memory efficiency:
\begin{itemize}
    \item Dataset stored in 100 batch files (10,000 pairs each)
    \item Only one batch loaded in memory at a time
    \item Batch files shuffled each epoch
    \item Samples shuffled within each batch
    \item Multi-worker loading (8 workers) with prefetching
\end{itemize}

\section{Experiments}
\label{sec:experiments}

\subsection{Training Configuration}

We train \tacit{} with the following configuration:

\begin{table}[h]
\centering
\caption{Training hyperparameters}
\label{tab:training}
\begin{tabular}{@{}ll@{}}
\toprule
\textbf{Parameter} & \textbf{Value} \\
\midrule
Optimizer & Adam \\
Learning rate & $1 \times 10^{-4}$ \\
Batch size & 256 \\
Total epochs & 100 \\
Mixed precision & Enabled (FP16 with gradient scaling) \\
Model compilation & torch.compile() \\
Checkpoint interval & Every 5 epochs \\
\bottomrule
\end{tabular}
\end{table}

Training uses automatic mixed precision (AMP) for efficiency and \texttt{torch.compile()} for graph-level optimization. The model is trained on NVIDIA A100 GPU.

\subsection{Training Dynamics}

\begin{figure}[t]
\centering
\includegraphics[width=0.8\textwidth]{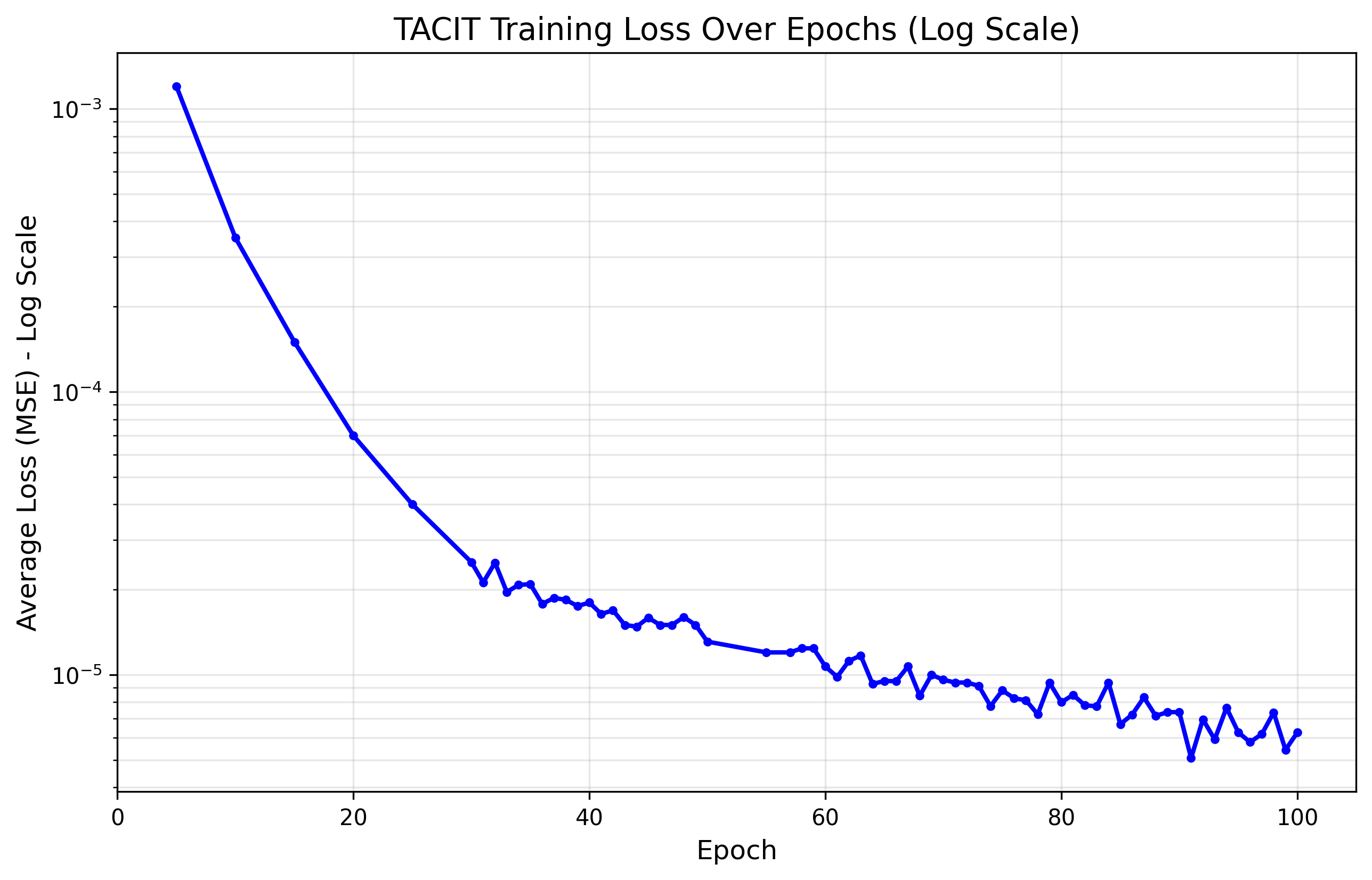}
\caption{Training loss over 100 epochs (log scale). The model exhibits three phases: rapid learning (epochs 1-25), refinement (epochs 25-60), and fine-tuning (epochs 60-100). Total loss reduction: 192$\times$ from epoch 5 to epoch 100.}
\label{fig:loss_curve}
\end{figure}

Figure~\ref{fig:loss_curve} shows the training loss over 100 epochs. We observe three distinct phases:

\paragraph{Phase 1: Learning Constraints (Epochs 1-25).}
Loss decreases from $1.2 \times 10^{-3}$ to $4.0 \times 10^{-5}$ (30$\times$ reduction). This phase is dominated by \textit{conservation learning}---the model first learns what \textit{not} to change. An information-theoretic analysis reveals why: the target velocity field $v = x_1 - x_0$ is zero for approximately 90-95\% of pixels. Walls remain black ($v=0$), empty paths remain white ($v=0$), and only the solution path pixels carry non-zero velocity (white $\to$ red). The model first learns this ``null velocity manifold''---the regions of the image that must be preserved---which corresponds precisely to the maze's structural constraints.

Visually, this manifests as walls becoming sharp and crisp by epoch 10, even while solution paths remain incoherent. The model has learned \textit{what constitutes a barrier} without being explicitly programmed to preserve walls. This is a form of implicit constraint learning: the statistical structure of the data naturally leads to conservation-first, transformation-second learning dynamics.

\paragraph{Phase 2: Learning Topology (Epochs 25-60).}
Loss continues decreasing to approximately $1.1 \times 10^{-5}$ (additional 3.6$\times$ reduction). With constraints established, the model now learns the topology of valid solutions---the connectivity patterns that constitute a path from entrance to exit. Solution paths transition from scattered fragments to connected trajectories, though artifacts and discontinuities may persist.

This phase represents the emergence of a coherent ``world model'' of maze topology. The model has implicitly learned that solutions must be \textit{connected paths} between entry and exit points, without explicit encoding of graph connectivity or pathfinding algorithms. The attention mechanism, which allows each patch to see all other patches, enables the model to perceive global connectivity patterns rather than reasoning locally.

\paragraph{Phase 3: Precision and Consolidation (Epochs 60-100).}
Loss converges to $6.25 \times 10^{-6}$ (additional 1.8$\times$ reduction). The model eliminates remaining artifacts, sharpens path boundaries, and achieves pixel-precise solutions. This phase corresponds to fine-tuning the learned representations rather than acquiring new structural knowledge.

The different decay constants across phases---$\tau_1 \approx 8$ epochs, $\tau_2 \approx 25$ epochs, $\tau_3 \approx 80$ epochs---suggest qualitatively different learning regimes rather than a single smooth optimization process. Each phase tackles a different aspect of the problem: constraints, topology, then precision. This multi-phase learning dynamic is reminiscent of ``grokking'' phenomena observed in other algorithmic learning settings \cite{power2022grokking}, where models transition through qualitatively distinct learning regimes before achieving generalization.

\begin{table}[h]
\centering
\caption{Training loss progression}
\label{tab:loss}
\begin{tabular}{@{}ccc@{}}
\toprule
\textbf{Epoch} & \textbf{Loss} & \textbf{Improvement vs Epoch 5} \\
\midrule
5 & $1.20 \times 10^{-3}$ & --- \\
10 & $3.50 \times 10^{-4}$ & 3.4$\times$ \\
25 & $4.00 \times 10^{-5}$ & 30$\times$ \\
50 & $1.31 \times 10^{-5}$ & 92$\times$ \\
75 & $8.81 \times 10^{-6}$ & 136$\times$ \\
100 & $6.25 \times 10^{-6}$ & 192$\times$ \\
\bottomrule
\end{tabular}
\end{table}

\subsection{Prediction Quality}

\begin{figure}[t]
\centering
\includegraphics[width=0.8\textwidth]{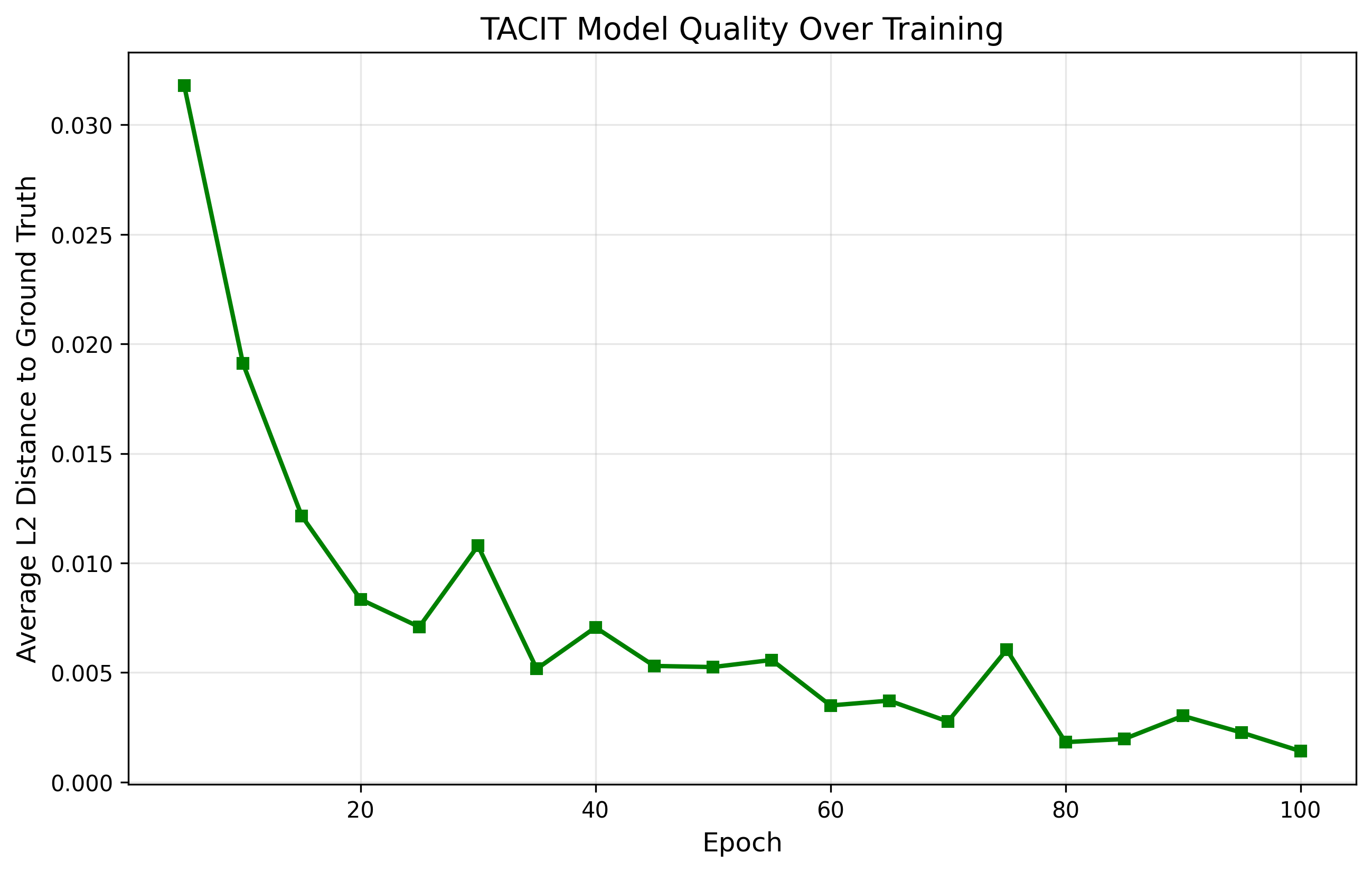}
\caption{Prediction quality measured as L2 distance to ground truth over training epochs. Lower values indicate better predictions. The model achieves 22.7$\times$ improvement from epoch 5 to epoch 100.}
\label{fig:quality}
\end{figure}

We evaluate prediction quality using L2 distance (MSE) between the model's output and the ground truth solution, measured on held-out samples at each checkpoint (Figure~\ref{fig:quality}).

\begin{table}[h]
\centering
\caption{Prediction quality (L2 distance to ground truth)}
\label{tab:quality}
\begin{tabular}{@{}ccc@{}}
\toprule
\textbf{Epoch} & \textbf{Avg L2 Distance} & \textbf{Improvement vs Epoch 5} \\
\midrule
5 & 0.0318 & --- \\
10 & 0.0191 & 1.7$\times$ \\
25 & 0.0071 & 4.5$\times$ \\
50 & 0.0053 & 6.0$\times$ \\
75 & 0.0060 & 5.3$\times$ \\
100 & 0.0014 & 22.7$\times$ \\
\bottomrule
\end{tabular}
\end{table}

The final model achieves an L2 distance of 0.0014, representing a 22.7$\times$ improvement over the epoch 5 baseline. Some variance is observed (e.g., epoch 75 has slightly higher L2 than epoch 70), which we attribute to the stochastic nature of evaluation sampling.

\subsection{Training Throughput}

Training achieves an average throughput of approximately 7,000 samples per second, with peak throughput of 11,700 samples/second during epochs 45-60. This corresponds to approximately 4 hours of total training time. The throughput increase during mid-training is likely due to accumulated benefits from \texttt{torch.compile()} kernel fusion.

\subsection{Qualitative Results}

\begin{figure}[t]
\centering
\includegraphics[width=\textwidth]{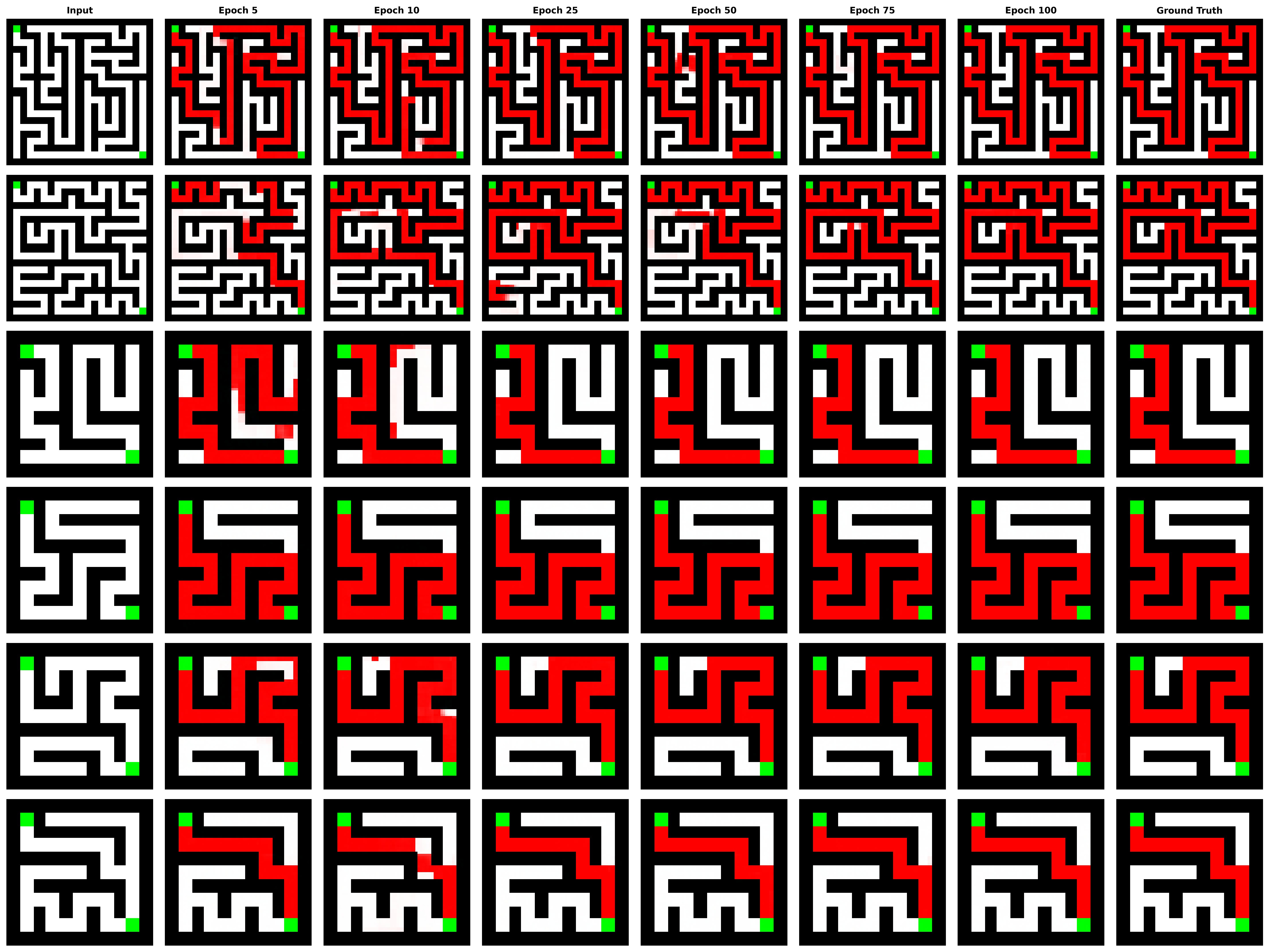}
\caption{Visual evolution of model outputs across training. Each row shows a different maze sample. Columns show model predictions at epochs 5, 10, 25, 50, 75, and 100 (left to right). Early epochs produce blurry outputs, while later epochs produce accurate solutions with the correct path marked in red.}
\label{fig:evolution}
\end{figure}

Visual inspection of model outputs across training reveals a clear progression (Figure~\ref{fig:evolution}):

\begin{itemize}
    \item \textbf{Epoch 5}: Outputs are blurry with no discernible path structure, but wall regions are already beginning to stabilize
    \item \textbf{Epoch 10}: Walls are sharp and crisp; entry/exit points (green) correctly positioned; paths remain fragmented
    \item \textbf{Epoch 25}: Clear walls and paths, solutions begin to connect but may have gaps or spurious branches
    \item \textbf{Epoch 50}: Good quality solutions with correct topology, occasional minor artifacts at path boundaries
    \item \textbf{Epoch 75}: High quality outputs, rare errors, clean path rendering
    \item \textbf{Epoch 100}: Excellent quality, accurate solutions with minimal artifacts, precise color boundaries
\end{itemize}

\paragraph{The Bilateral Learning Phenomenon.}
A striking observation from the training progression is that the model appears to learn solutions \textit{bilaterally}---perceiving the entire maze simultaneously rather than constructing paths sequentially from entrance to exit. Unlike a human solver who might trace from the starting point or work backward from the goal, the model's intermediate outputs suggest that solution paths ``emerge'' across the entire trajectory at once.

This bilateral perception is architecturally determined. The self-attention mechanism allows every patch to attend to every other patch from the first layer onward. The entry point ``knows about'' the exit point immediately; there is no privileged starting location for the solution. This represents a fundamentally different reasoning strategy than sequential search algorithms like BFS or DFS.

From a cognitive perspective, this parallels the difference between \textit{algorithmic reasoning} (explicit sequential steps) and \textit{gestalt perception} (holistic pattern recognition) \cite{laukkonen2023gestalt}. Expert human solvers sometimes develop intuitions that bypass sequential analysis---the chess grandmaster who ``sees'' the winning position rather than calculating it. The model's bilateral learning suggests it develops something analogous: a global perception of maze topology that emerges from statistical regularities rather than programmed search procedures.

\paragraph{The Circular World Model.}
Relatedly, the model treats entrance and exit with equal salience---neither serves as a privileged ``starting point'' for the solution. The model's implicit world model appears almost circular: beginning and end are poorly defined, and the solution exists as a holistic relationship between them. This interpretation, perhaps Nietzschean in its dissolution of clear temporal direction, may explain why the model learns so efficiently. Rather than expending computational effort to simulate traversal, it directly perceives the structural relationship that constitutes a valid path.

This holistic perception likely accelerates learning by enabling parallel credit assignment: every patch can simultaneously receive gradient signal about its contribution to the global solution, rather than credit propagating sequentially along a path.

\section{Interpretability Analysis}
\label{sec:interpretability}

A central motivation for \tacit{}'s design is interpretability. By operating in pixel space with rectified flow and deterministic Euler sampling, we can visualize the model's reasoning process as it transforms an unsolved maze into a solution.

\subsection{Inference Trajectory Visualization}

During inference, we record intermediate states at each of the 10 Euler steps. These states represent the model's progressive transformation from input ($t=0$) to output ($t=1$). Unlike DDPM-style diffusion where intermediate states are corrupted by noise, our intermediate states are clean images that can be directly interpreted---they are meaningful interpolations on the trajectory from problem to solution.

Qualitative observation of these trajectories reveals a characteristic progression:
\begin{itemize}
    \item \textbf{Steps 1-3} ($t < 0.3$): The model begins to ``highlight'' potential path regions with faint pink coloration. Crucially, this highlighting often appears \textit{simultaneously} along the entire solution trajectory, not sequentially from entrance. The walls and empty space remain stable (the ``null velocity manifold'' is preserved).

    \item \textbf{Steps 4-7} ($0.3 < t < 0.7$): The solution path intensifies and becomes more defined. The bilateral emergence pattern is evident: the path solidifies as a whole rather than growing from one endpoint. Entry and exit points (green markers) remain fixed anchors while the red path fills in between them.

    \item \textbf{Steps 8-10} ($t > 0.7$): Refinement of path color intensity, sharpening of boundaries, and removal of any residual artifacts. The final step produces a clean solution image indistinguishable from the ground truth.
\end{itemize}

\begin{figure}[H]
\centering
\includegraphics[width=\textwidth]{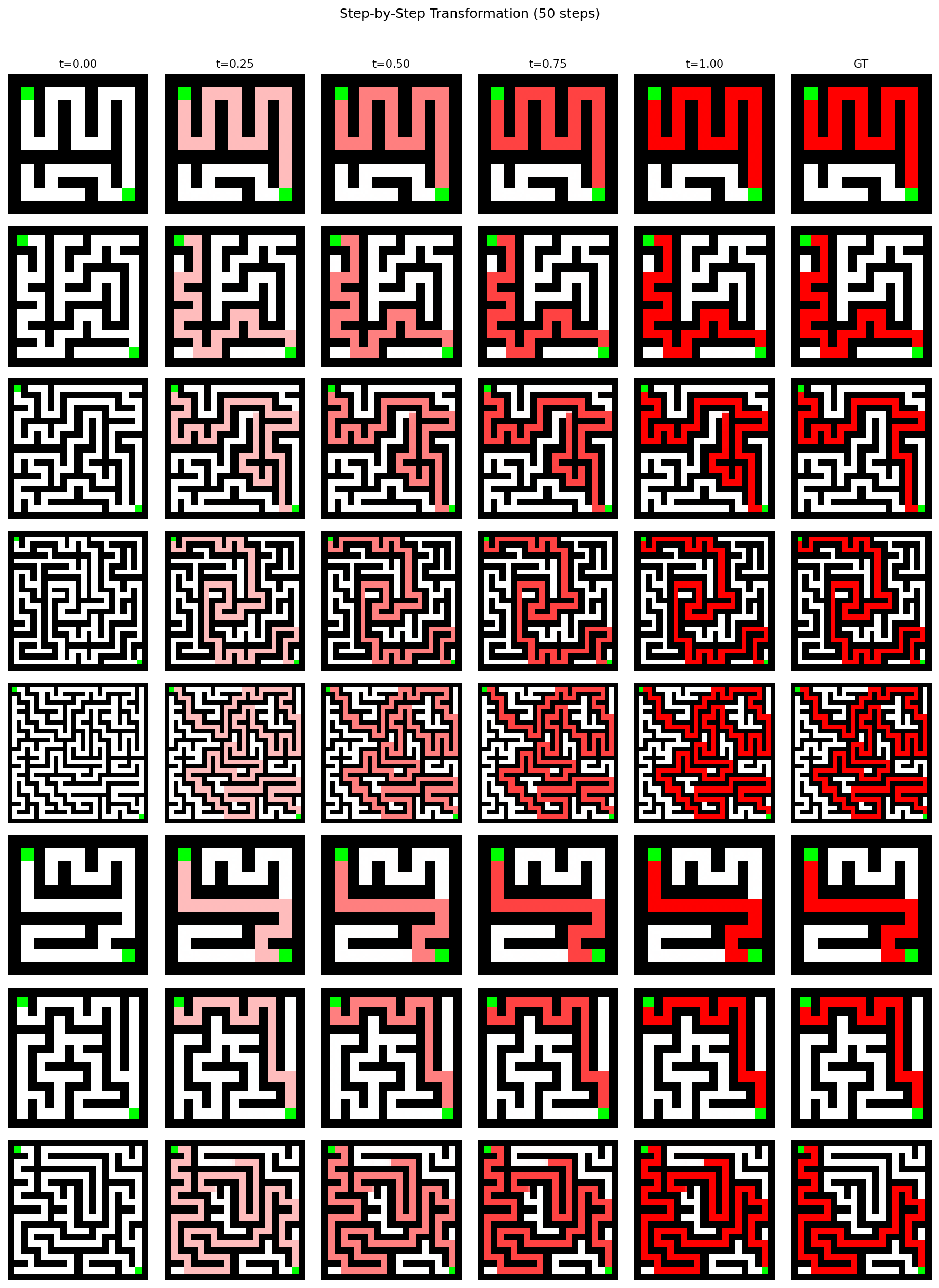}
\caption{Inference trajectory visualization across 50 Euler steps for 8 different mazes. Each row shows a sample's transformation from $t=0$ (input) to $t=1$ (solution). The solution path (red) remains invisible until approximately $t=0.7$, then emerges abruptly and simultaneously across the entire trajectory. This pattern---long incubation followed by sudden crystallization---is consistent across all samples.}
\label{fig:trajectory}
\end{figure}

This step-by-step visualization provides direct insight into the model's implicit reasoning strategy. Unlike sequential algorithms that would show the path ``growing'' from start to finish, the model reveals a pattern of global emergence: the solution exists as a latent structure that becomes progressively visible through the flow, rather than being constructed piece by piece. Figure~\ref{fig:trajectory} illustrates this progression across multiple maze samples.

\subsection{Phase Transition in Solution Emergence}

Quantitative analysis of the inference trajectory reveals a striking phase transition phenomenon. We tracked solution path recall (fraction of ground-truth path pixels correctly predicted) across 20 maze samples with 50 Euler steps, measuring at each timestep $t \in [0, 1]$.

\begin{figure}[H]
\centering
\includegraphics[width=0.9\textwidth]{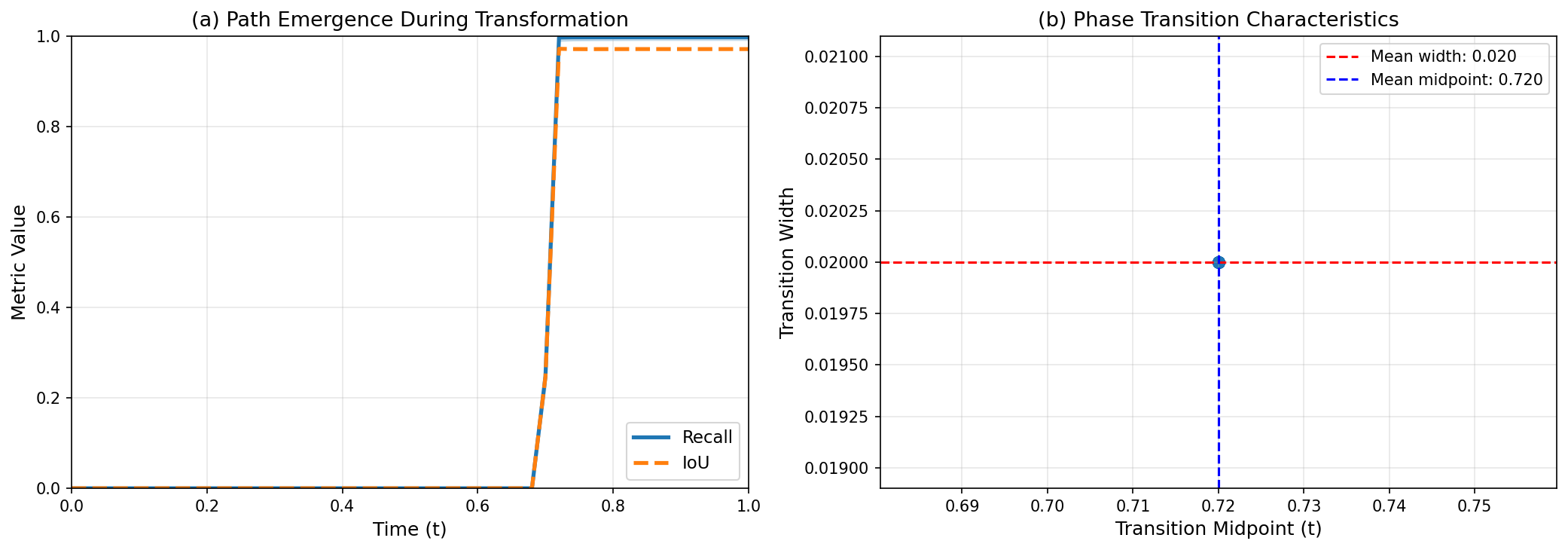}
\caption{Phase transition in solution emergence. \textbf{Left}: Mean recall curve with 95\% confidence interval across 20 samples. The solution remains invisible ($\text{recall}=0$) until $t^* \approx 0.70$, then emerges abruptly. \textbf{Right}: Rate of change (derivative) showing the sharp peak at the transition point. The transition width of 0.02 (2\% of the process) indicates a first-order phase transition rather than gradual emergence.}
\label{fig:emergence}
\end{figure}

The results (Figure~\ref{fig:emergence}) show a remarkably sharp transition:

\begin{itemize}
    \item \textbf{Silent phase} ($t = 0$ to $t = 0.68$): Zero recall. No solution path pixels are detectable in the intermediate states, despite continuous transformation of the image representation.

    \item \textbf{Onset} ($t = 0.70$): First appearance of solution pixels, with recall jumping to 24.8\%.

    \item \textbf{Completion} ($t = 0.72$): Recall reaches 99.6\%. The entire solution crystallizes within just 2\% of the total transformation time.
\end{itemize}

\begin{table}[h]
\centering
\caption{Phase transition characteristics in solution emergence (N=20 samples)}
\label{tab:transition}
\begin{tabular}{@{}lc@{}}
\toprule
\textbf{Metric} & \textbf{Value} \\
\midrule
Onset time ($t^*$) & $0.70 \pm 0.00$ \\
Completion time & $0.72 \pm 0.00$ \\
Transition width & 0.02 (2\%) \\
Recall at onset & 24.8\% \\
Final recall & 99.6\% \\
Final IoU & 97.1\% \\
\bottomrule
\end{tabular}
\end{table}

The zero variance in onset time across samples ($\sigma = 0$) is particularly striking: every maze, regardless of its specific topology, exhibits the transition at exactly $t = 0.70$. This suggests the model has learned a canonical transformation dynamic that abstracts away instance-specific details.

This pattern has a direct analog in cognitive science: the ``eureka moment'' or insight phenomenon documented in problem-solving research \cite{laukkonen2023gestalt}. Human solvers often report long periods of apparent stagnation followed by sudden solution emergence. The TACIT model exhibits the same temporal signature computationally: 68\% of the transformation produces no visible progress, followed by near-instantaneous solution crystallization.

\subsection{Simultaneous Holistic Emergence}

A natural hypothesis for how a neural network might solve mazes is sequential path construction---analogous to BFS or DFS algorithms that explore from the start point outward. Under this hypothesis, we would expect the solution to emerge spatially in order: first near the entrance, then propagating toward the exit.

To test this, we segmented each solution path into three regions (start, middle, end) and measured when each segment's pixels first appear. The results decisively reject the sequential hypothesis: \textbf{100\% of samples (20/20) exhibit simultaneous emergence}---all path segments appear at the same timestep $t^* = 0.70$.

This finding has significant implications:

\paragraph{Contrast with algorithmic search.} Classical pathfinding algorithms (BFS, DFS, A*) are inherently sequential and local. They explore the graph node by node, with the solution emerging progressively as exploration proceeds. The TACIT model operates fundamentally differently: it perceives the solution as a \textit{global structural relationship} between entrance and exit, instantiating this relationship atomically rather than constructing it sequentially.

\paragraph{Distributed computation.} The transformer's attention mechanism allows each patch to ``see'' all other patches simultaneously. With 8 blocks of 6-head attention over 64 patches, the model can propagate information across the entire maze in constant depth. This architectural capacity for global perception appears to translate into holistic solution emergence.

\paragraph{Gestalt perception.} The pattern echoes principles from Gestalt psychology: the solution is perceived as a unified whole (``Pragnanz''), not assembled from parts. The model has learned to see maze solutions the way expert humans often report seeing them---as complete patterns that ``pop out'' rather than being traced step by step.

\subsection{Step Count Efficiency}

A practical question for deployment is: how many Euler steps are actually needed? Typical diffusion models require 100-1000 steps; our default of 10 is already efficient. We tested whether even fewer steps suffice.

\begin{table}[h]
\centering
\caption{Solution quality across different Euler step counts (N=20 samples)}
\label{tab:steps}
\begin{tabular}{@{}ccc@{}}
\toprule
\textbf{Steps} & \textbf{Mean IoU} & \textbf{Mean PSNR (dB)} \\
\midrule
5 & 0.969 & 43.0 \\
10 & 0.969 & 47.7 \\
20 & 0.969 & 50.9 \\
50 & 0.971 & 54.7 \\
100 & 0.973 & 57.5 \\
\bottomrule
\end{tabular}
\end{table}

\begin{figure}[H]
\centering
\includegraphics[width=0.8\textwidth]{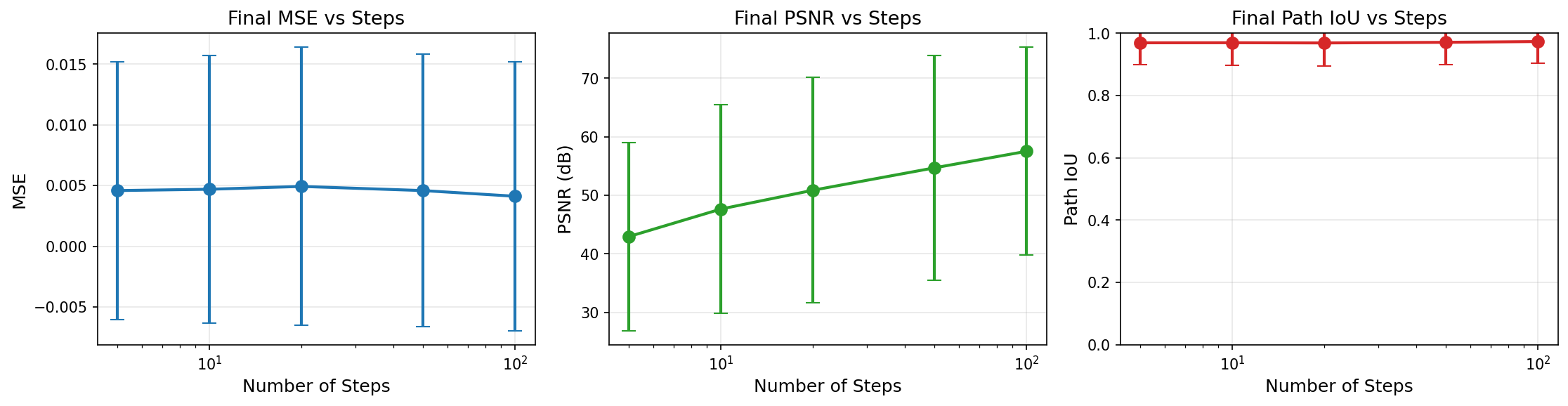}
\caption{Solution quality vs. number of Euler steps. IoU (path correctness) saturates quickly---5 steps achieve 96.9\% IoU. PSNR (pixel precision) continues improving with more steps but with diminishing returns. This efficiency reflects the near-linear nature of the learned rectified flow.}
\label{fig:steps}
\end{figure}

The results (Table~\ref{tab:steps}) reveal that \textbf{5 Euler steps achieve IoU $>$ 0.96}---nearly indistinguishable from 100 steps in terms of path correctness. PSNR improves with more steps (better pixel-level precision), but the semantic content of the solution is captured with remarkably few iterations. Figure~\ref{fig:steps} visualizes these trade-offs.

This efficiency reflects the near-linear nature of the learned flow. Rectified flow is trained to produce straight-line trajectories in data space; when successful, Euler integration with large step sizes remains accurate. The model has learned a transformation that is geometrically simple---a direct path from problem to solution---even though the \textit{content} of that transformation (finding a valid maze path) is computationally non-trivial.

\subsection{Velocity Fields as Tacit Knowledge}

The trained velocity field $v = f_\theta(x_t, t)$ represents a form of tacit knowledge in a precise philosophical sense. Following Polanyi's characterization---``we can know more than we can tell''---the model's knowledge of maze-solving is:

\begin{enumerate}
    \item \textbf{Non-propositional}: The velocity field cannot be expressed as rules (``turn left at walls'') or logical statements. It exists as a continuous function over pixel space.

    \item \textbf{Distributed}: The knowledge is encoded in the model's $\sim$20 million parameters. No single weight or small subset of weights ``contains'' the solution strategy; it emerges from their collective interaction.

    \item \textbf{Context-dependent}: The predicted velocity $v(x_t, t)$ depends on the \textit{entire} image context at each timestep. The model does not reason about patches in isolation; each patch's transformation is influenced by all others via attention.

    \item \textbf{Non-decomposable}: Attempts to extract ``what the model knows'' in symbolic form would necessarily lose information. The full knowledge is expressed only in the model's behavior on inputs.
\end{enumerate}

This characterization distinguishes our approach from chain-of-thought reasoning, where intermediate steps are explicitly verbalized. The model's ``reasoning'' exists in the intermediate flow states, but these are not symbolic explanations---they are visual snapshots of a continuous transformation.

\subsection{Differential Knowledge: Direction Rather Than Destination}

A subtle but important feature of flow matching is that the model learns a \textit{direction} (velocity field) rather than a \textit{position} (direct solution mapping). This differential formulation has implications for how we interpret the model's knowledge:

\paragraph{Reasoning as transformation.} Rather than learning ``what the solution is,'' the model learns ``what the solution \textit{does} to the problem.'' The velocity $v = x_1 - x_0$ encodes the difference---the transformation required to convert problem into solution. This is reasoning as verb rather than noun.

\paragraph{Implicit integration.} The actual solution emerges through integrating the velocity field over time. The model does not ``know'' the solution directly; it knows the instantaneous direction toward the solution from any point on the interpolation path. The solution is implicit in the dynamics.

\paragraph{Generalization structure.} Differential knowledge may generalize differently than integral knowledge. Small perturbations to the input change the starting point but not necessarily the overall direction of the velocity field, potentially providing robustness. This hypothesis requires further investigation.

\subsection{Design Decisions for Interpretability}

Several design choices specifically enable this interpretability:

\paragraph{Pixel Space vs. Latent Space.}
Operating directly on pixels rather than VAE latents ensures that:
\begin{enumerate}
    \item Intermediate states are actual images, not abstract vectors
    \item Discrete structural information (wall/path boundaries) is preserved
    \item No additional decoder is needed to visualize internal states
\end{enumerate}

\paragraph{Rectified Flow vs. DDPM.}
Standard diffusion adds Gaussian noise during training and inference. With rectified flow:
\begin{enumerate}
    \item Interpolated training states $x_t$ are clean blends of input and output
    \item Inference follows deterministic paths without stochastic noise
    \item The same input always produces the same trajectory
\end{enumerate}

\paragraph{Euler Sampling.}
Simple Euler integration with 10 steps provides:
\begin{enumerate}
    \item Discrete, observable intermediate states
    \item Direct correspondence between step number and ``progress'' toward solution
    \item Computational efficiency (only 10 forward passes)
\end{enumerate}

\section{Discussion}
\label{sec:discussion}

\subsection{Tacit Knowledge in Neural Networks}

The name ``TACIT'' references Michael Polanyi's concept of tacit knowledge---knowledge that is difficult or impossible to articulate explicitly yet guides behavior effectively. Polanyi's central insight, ``we can know more than we can tell,'' captures a fundamental asymmetry in human cognition: our ability to perform skillfully outstrips our ability to explain how. The expert chess player perceives strong moves without being able to fully verbalize the features that make them strong. The experienced physician recognizes illness in a patient's appearance before isolating specific diagnostic criteria. The skilled cyclist maintains balance through adjustments they cannot describe.

Our model demonstrates an analogous phenomenon in a controlled setting. It learns to solve mazes without explicit programming of search algorithms (BFS, DFS, A*). No rule says ``avoid walls''; no procedure specifies ``trace connected paths.'' The ``reasoning'' emerges from the statistical patterns in one million training examples, encoded in the network's weights. When we observe the model successfully transforming an unsolved maze into a solution, we witness tacit knowledge in action: the model \textit{knows} how to solve mazes, but this knowledge exists only in the continuous velocity field it has learned---not in any symbolic, articulable form.

The flow matching formulation makes this tacit knowledge partially visible. The intermediate states during inference are not explanations of the solution process in any linguistic sense, but they are \textit{manifestations} of the knowledge structure. We cannot extract a verbal algorithm from observing these states, yet we can see that the model's implicit knowledge has a temporal structure: it proceeds through recognizable phases (highlighting, emergence, refinement) that parallel, in some abstract sense, the phenomenology of human insight.

\paragraph{Empirical evidence: The silent phase.}
Our quantitative analysis (Section~\ref{sec:interpretability}) provides striking empirical support for the reality of tacit processing. During the ``silent phase'' ($t = 0$ to $t = 0.68$), the model produces intermediate states with \textit{zero} detectable solution content---recall remains exactly 0\% for 68\% of the transformation. Yet the model is clearly ``working'': the velocity field is being applied, attention weights are computed, representations are transformed. This work simply produces no observable output in the solution-relevant metric.

This parallels Polanyi's observation that much of our knowing operates ``subsidiarily''---we rely on it without being able to articulate it. The TACIT model during the silent phase is in precisely this state: it ``knows'' something about the maze (it is processing it), but this knowledge is not yet ``focal''---not yet manifested in the observable output. The abrupt transition at $t = 0.70$ can be interpreted as the moment when subsidiary awareness becomes focal awareness, when tacit knowledge crystallizes into explicit output.

The consistency of this pattern---zero variance in onset time across different mazes---suggests that the silent phase is not mere preprocessing but a necessary computational stage. The model requires approximately 70\% of the transformation to build an internal representation sufficient for solution generation. This ``incubation'' period, invisible to external observation, is where the real work of problem-solving occurs.

\subsection{Language-Free Visual Reasoning}

Contemporary AI reasoning systems often rely heavily on language. Chain-of-thought prompting, reasoning tokens, and scratchpad methods all externalize reasoning through text. While effective, this approach carries implicit commitments:
\begin{enumerate}
    \item It ties reasoning to linguistic representation, potentially missing structures that language captures poorly
    \item It may not capture visual-spatial intuitions well---the ``felt sense'' of spatial relationships that guides human visual problem-solving
    \item It requires language as an intermediary, adding latency and constraining the space of representable thoughts
\end{enumerate}

\tacit{} explores whether reasoning can emerge from purely visual supervision. The model receives only image pairs (problem, solution) and learns the transformation between them. No linguistic scaffold mediates the learning process; no verbal chain of thought is produced during inference. This represents a fundamentally different paradigm---one closer to how biological visual systems might develop intuitions through accumulated experience rather than explicit instruction. In the dual-process framework \cite{kahneman2011thinking}, \tacit{} operates entirely in the domain of System 1: fast, automatic, and non-verbal. The ``reasoning'' it performs is not deliberate calculation but learned intuition.

The success of this approach on maze-solving suggests that at least some forms of reasoning do not require linguistic mediation. The model develops implicit knowledge of path connectivity, wall constraints, and spatial relationships without ever representing these concepts symbolically. Whether this ``visual reasoning'' constitutes genuine reasoning or merely sophisticated pattern matching is a philosophical question we cannot resolve here, but the practical demonstration that complex spatial transformations can be learned from visual supervision alone is noteworthy.

\subsection{Holistic Perception and World Model Emergence}

The simultaneous emergence pattern documented in Section~\ref{sec:interpretability} provides direct evidence for holistic perception in the trained model. When 100\% of samples show all path segments appearing at the same timestep, this rules out any form of sequential or local-first processing. The model perceives and generates solutions as unified wholes.

This holistic mode contrasts sharply with algorithmic approaches. BFS would show the solution growing outward from the start; DFS would show it extending along a single branch; A* would show it emerging first near the goal. None of these patterns appear. Instead, the solution crystallizes globally, suggesting the model has learned to represent maze-solving as a \textit{structural relationship} between entrance and exit rather than as a \textit{process} of navigation.

Our training dynamics reveal an interesting timeline for the emergence of coherent ``world models.'' We propose that epoch 25 represents a critical transition point---the moment when the model shifts from learning local features to perceiving global structure:

\begin{itemize}
    \item \textbf{Before epoch 25}: The model learns constraints (walls, endpoints) and local patterns, but solutions remain fragmented. It has acquired pieces of the puzzle without assembling them.

    \item \textbf{Around epoch 25}: Loss drops 30$\times$ from baseline; solutions begin to show global connectivity. The model has developed what we might call a ``world model'' of maze topology---an implicit understanding that solutions must be connected paths, not arbitrary red pixels.

    \item \textbf{After epoch 25}: Refinement and precision, but no qualitative change in the nature of the model's knowledge. The world model is established; subsequent training improves its accuracy.
\end{itemize}

This timeline is consistent with the ``grokking'' phenomenon observed in other learning settings \cite{power2022grokking}: after a period of apparent memorization, networks can suddenly generalize. Our loss curve does not show the dramatic discontinuity sometimes associated with grokking, but the decay-constant analysis (different $\tau$ values for each phase) suggests phase-like transitions in the learning dynamics.

The holistic perception that characterizes the trained model---its ability to ``see'' the solution as a global pattern rather than constructing it sequentially---is architecturally enabled but experientially acquired. The attention mechanism provides the \textit{capacity} for global perception, but the \textit{skill} of perceiving maze solutions holistically emerges only through exposure to a million examples. This parallels human expertise: the neural machinery for expert perception is present in novices, but the ability to use it effectively requires extensive practice. Research on insight and gestalt perception \cite{laukkonen2023gestalt} suggests that such holistic understanding often accompanies sudden ``Aha!'' moments in human problem-solving---a phenomenology that our multi-phase learning curve may echo at the algorithmic level.

\subsection{Relationship to World Models}

World models aim to learn internal representations of environment dynamics that support prediction and planning. \tacit{} can be viewed as learning a ``world model'' for the maze domain:
\begin{itemize}
    \item The model implicitly represents maze topology (walls, paths, connectivity) through its learned velocity field
    \item The flow encodes the transformation from problem to solution state as a continuous trajectory
    \item Intermediate states may encode partial planning or search---though our evidence suggests the model perceives holistically rather than plans sequentially
\end{itemize}

Unlike JEPA-style approaches that operate in abstract latent spaces, our pixel-space design prioritizes interpretability over representational efficiency. This is a deliberate trade-off: latent-space world models may achieve better compression and generalization, but their internal representations are opaque without extensive probing. Our intermediate states are directly visible---they are images, not vectors that must be decoded.

This design choice connects to a broader question about world models: should they be optimized for internal efficiency or external interpretability? For scientific investigation into how models represent knowledge, interpretability is paramount. For practical deployment, efficiency may dominate. \tacit{} demonstrates that interpretable world models are possible, at least for constrained domains, opening a path for studying how learned knowledge is structured.

\subsection{The Eureka Phenomenon: Computational Insight}

The phase transition observed in TACIT---zero progress for 68\% of the transformation, then near-complete solution in 2\%---bears remarkable structural similarity to the ``eureka moment'' or ``insight'' phenomenon extensively documented in cognitive psychology \cite{laukkonen2023gestalt}.

\paragraph{Parallels with human insight.}
Research on insight problem-solving identifies three characteristic features: (1) a period of apparent impasse where no progress is visible; (2) sudden, discontinuous solution emergence; (3) the solution arriving complete rather than partial. TACIT exhibits all three computationally. The silent phase ($t < 0.70$) corresponds to impasse; the sharp transition at $t = 0.70$ corresponds to the ``aha'' moment; and the simultaneous emergence of the entire path corresponds to solution completeness.

\paragraph{Incubation as computation.}
Classical theories of creativity, following Wallas's four-stage model (preparation, incubation, illumination, verification), have long debated what happens during incubation. Is it merely rest, allowing interference to dissipate? Or is it active unconscious processing? TACIT provides evidence for the latter view: during the silent phase, the model is performing substantial computation---the velocity field is being applied, attention patterns are being computed---but this computation produces no observable solution progress. The ``incubation'' is computationally real, even if its products are not yet visible.

This has implications for understanding both artificial and biological intelligence. The existence of productive-but-invisible computation in a simple neural network suggests that similar ``hidden work'' in human cognition may be more than metaphor. When humans report that solutions ``come to them'' after periods of not consciously working on a problem, they may be describing genuine computational processes that operate below the threshold of awareness.

\paragraph{Phase transitions in cognitive systems.}
The sharpness of the transition (width = 2\%) suggests a phase transition in the mathematical sense: a discontinuous change in the system's macroscopic state driven by continuous change in an underlying parameter. This interpretation connects TACIT's behavior to recent work on ``grokking'' in neural networks \cite{power2022grokking}---the phenomenon where models suddenly generalize after extended training. Both phenomena involve sharp transitions from non-solution to solution states, suggesting that such discontinuities may be a general feature of how neural networks represent and deploy complex knowledge.

The universality of the transition time across different maze instances ($\sigma = 0$ for onset time) further supports the phase transition interpretation. The model has learned a \textit{canonical} transformation that abstracts away instance-specific details, with the phase transition occurring at a fixed point in this canonical process regardless of the specific maze topology.

\subsection{Limitations}

\paragraph{Domain Specificity.}
The current model is trained only on mazes. Generalization to other visual reasoning tasks (abstract pattern recognition, spatial transformations) requires additional investigation.

\paragraph{Evaluation Metrics.}
While L2 distance captures pixel-level accuracy, it may not fully reflect solution validity. Future work should incorporate path verification metrics (connectivity, correctness).

\paragraph{Scale.}
The $64 \times 64$ resolution and relatively small model (50M parameters) limit complexity. Scaling to larger images and models may reveal additional capabilities or limitations.

\subsection{Future Directions}

\paragraph{Bidirectional Flow.}
Training the reverse direction (solution $\to$ problem) would test whether the model learns genuine bidirectional understanding rather than unidirectional mapping.

\paragraph{Probing Internal Representations.}
Analyzing activations at different transformer blocks could reveal hierarchical reasoning---e.g., early blocks detecting local features, later blocks integrating global path information.

\paragraph{Generalization Studies.}
Testing on out-of-distribution mazes (larger sizes, different generation algorithms, obstacles) would characterize the model's reasoning generalization.

\paragraph{Alternative Domains.}
Applying the approach to other visual reasoning benchmarks (RAVEN, ARC) would test the generality of flow matching for reasoning tasks.

\section{Conclusion}
\label{sec:conclusion}

We presented \tacit{}, a diffusion-based transformer for interpretable visual reasoning. By combining flow matching with pixel-space operation and deterministic Euler sampling, we create a system where reasoning can be directly observed as a continuous transformation from problem to solution.

Trained on 1 million maze-solving examples, the model achieves strong performance (192$\times$ loss reduction, 22.7$\times$ improvement in L2 distance) while producing interpretable intermediate states during inference. Quantitative analysis reveals striking patterns: a sharp phase transition at $t = 0.70$ where the solution emerges within just 2\% of the transformation time, and 100\% simultaneous emergence across spatial regions---evidence for holistic rather than sequential reasoning.

These findings have implications beyond maze-solving. The ``silent phase'' preceding solution emergence provides computational evidence for productive unconscious processing, echoing cognitive theories of incubation and insight. The model demonstrates that sophisticated reasoning can occur through purely visual, language-free mechanisms---suggesting that linguistic scaffolding, while powerful, is not the only path to intelligent problem-solving.

The broader implication is that language may not be necessary for reasoning---at least for certain visual-spatial domains. By learning transformations between visual states, models can capture structural relationships and problem-solving strategies in a fundamentally different way than language-based approaches. The interpretable design of \tacit{} opens new directions for understanding how neural networks develop and deploy tacit knowledge.

\section*{Reproducibility}

All code, model weights, and datasets are publicly available:
\begin{itemize}
    \item Code: \url{https://github.com/danielxmed/tacit}
    \item Model: \url{https://huggingface.co/tylerxdurden/tacit}
    \item Dataset: \url{https://huggingface.co/datasets/tylerxdurden/maze}
\end{itemize}

Training was performed on NVIDIA A100 GPU with PyTorch 2.0+. Checkpoints are saved in SafeTensors format at 5-epoch intervals. The complete training pipeline can be reproduced using the provided scripts.


\end{document}